\documentclass[11pt]{article}

\usepackage[a4paper,margin=2.4cm]{geometry}
\usepackage[T1]{fontenc}
\usepackage{lmodern}
\usepackage[english]{babel}
\usepackage{amsmath,amssymb,amsthm}
\usepackage{booktabs}
\usepackage{array}
\usepackage{graphicx}
\usepackage{microtype}
\usepackage{xcolor}
\usepackage{listings}
\usepackage[round,authoryear]{natbib}
\usepackage{hyperref}
\hypersetup{colorlinks=true,linkcolor=blue!50!black,citecolor=blue!50!black,urlcolor=blue!50!black}

\title{Your SaaS Is an Insurance Product:\\
A Modeling Framework}
\author{%
  Caio Gomes\\
  Magalu\\
  \href{mailto:gomes.caio@luizalabs.com}{\texttt{gomes.caio@luizalabs.com}}%
}
\date{May 2026}

\begin{document}
\maketitle

\begin{abstract}
Capped-usage SaaS products---LLM subscriptions such as Claude Code and ChatGPT, cloud platforms such as Vercel and Cloudflare Workers, corporate benefit platforms, identity-verification services with liability transfer---share a structural signature with insurance products: a fixed premium decoupled from realized consumption, stochastic per-user demand with heavy-tailed severity, a non-fungible cap that resets on a fixed schedule, and a portfolio-level exposure that requires reserve adequacy under tail risk. We argue that this is not an analogy. It is the same operational problem actuarial science has been tooled for decades to address, restated with new dependent variables (tokens, bandwidth bytes, function-invocations, gym check-ins) in place of medical claims. This paper proposes a modeling framework for capped-usage SaaS pricing built from frequency--severity decomposition, premium calculation principles, and Monte Carlo reserve adequacy. We map the framework to publicly observable subscription tiers in two domains (LLM services and cloud platforms), ground it in canonical health-insurance economics \citep{arrow1963uncertainty,pauly1968moralhazard,manning1987rand,newhouse1993free,brotgoldberg2017deductible}, and demonstrate divergence from traditional unit economics through a worked example. The contribution is operational rather than theoretical: not a new theorem, but vocabulary and tools currently absent from cs.LG/stat.ML practice.
\end{abstract}

\section{Introduction}

Public reports of capped-usage AI subscriptions document an order-of-magnitude gap between the heaviest users' realized consumption at posted metered prices and their subscription revenue. A single documented incident on the Claude Code repository shows a Max-tier subscriber accruing roughly \$1{,}800 in API-equivalent charges over two days when a misconfiguration routed their usage through a metered key rather than the subscription.\footnote{GitHub issue \texttt{anthropics/claude-code\#37686}, documented 20--21 March 2026, \url{https://github.com/anthropics/claude-code/issues/37686}, accessed May 2026. The dollar figure is API-equivalent at posted token rates; this is an observable proxy for upper-bound seller exposure under public metered pricing, not the provider's marginal cost of inference (which is private and lower).} The subscriber stayed within contract; Anthropic stayed solvent. The arithmetic that closes the gap between subscription revenue and realized cost across the population is not pricing strategy. It is risk pooling. Light users subsidize heavy users, expected aggregate cost stays below total premium, and the firm holds a reserve against months when the heavy tail materializes more aggressively than expected. This is the operational identity of insurance---specifically, of a policy-limit insurance contract, in which the seller's payout is capped at a contractual ceiling regardless of realized consumption and the user absorbs anything above that ceiling.

Capped-usage SaaS---subscription tiers with a non-fungible monthly or weekly cap, after which service degrades, halts, or shifts to per-unit overage until the reset date---is the dominant commercial form for many software categories beyond LLMs. In LLM access alone, Anthropic introduced weekly limits to Claude Code in August 2025, repriced enterprise plans in November 2025, and tested removing Claude Code from the Pro tier in April 2026; OpenAI maintains parallel screening across ChatGPT Plus and Pro; GitHub Copilot has signaled a migration to usage-based pricing; Zhipu raised prices by 30\% in response. The same architecture appears in cloud platforms (Vercel Pro at \$20/month with bandwidth and function-invocation caps; Cloudflare Workers Paid; Fly.io's \$5/month plan), in database-as-a-service tiers (Supabase Pro, PlanetScale, Neon), in CI/CD products (GitHub Actions, CircleCI), and outside software in corporate benefit platforms (Wellhub) and identity-verification services with embedded liability transfer. The category exists, is large, and is regulated by the firm rather than by an external authority.

Current modeling practice in the engineering and data-science communities that build and operate these services treats pricing as a hybrid of traditional unit economics (cost per call multiplied by expected volume) and ad hoc heuristics for tier design (``is this user too cheap?''). We argue that a better-suited toolkit already exists in a discipline most modeling practitioners in this space have no contact with. Actuarial science has spent a century building machinery for closely related problems: frequency--severity decomposition of aggregate loss, expected-loss pricing with safety loading, Incurred-But-Not-Reported (IBNR) accounting for delayed claims, and reserve adequacy via Monte Carlo on the aggregate loss distribution. The behavioral component---what a rational user does as a non-fungible cap with reset approaches expiration---is structurally identical to the intra-year consumption acceleration documented in health economics when an insured patient reaches the out-of-pocket maximum: marginal price drops to zero, elective demand brought forward \citep{brotgoldberg2017deductible,einav2015response}.

This paper does not propose a new theorem. It proposes vocabulary and tools. We argue four points, each operational:

\begin{enumerate}
\item Capped-usage SaaS satisfies a structural definition of insurance products that does not depend on legal classification or regulatory treatment. The mathematics is the same regardless of whether the dependent variable is tokens, bandwidth bytes, function-invocations, or gym check-ins.
\item Frequency--severity decomposition with cap-aware censoring is the correct primitive for cost modeling, not naive expected-cost-times-volume.
\item Reserve dimensioning under VaR and TVaR on the aggregate per-period loss distribution gives a principled answer to questions currently solved by gut feel (``how much margin should we hold against a bad month?'').
\item The behavioral component, induced demand near cap reset, is identifiable and testable, and the actuarial literature has machinery for it.
\end{enumerate}

\paragraph{Framing: we model the seller's exposure.} This paper analyzes the seller's operational position under a capped-usage contract, not the buyer's. Whether the buyer is purchasing risk transfer is orthogonal to whether the seller's exposure has the structure of an insurance portfolio. We argue that the seller's structure is operationally that of an insurer running a portfolio of policies with a policy limit equal to $K_i$: fixed premium income, stochastic aggregate cost capped at $K_i$ per policy, tail risk over the portfolio, and a corresponding reserve adequacy requirement. The standard actuarial toolkit applies directly.

Policy-limit insurance is canonical, not exceptional. Dental and vision plans cap annual benefit at \$1{,}000 to \$3{,}000; auto liability policies cap per-accident payout; property policies cap total payout at the stated policy limit; reinsurance excess-of-loss treaties retain exposure above a contractual ceiling on the cedent. In each case the insured absorbs the tail above the cap, the insurer's exposure is the unceded portion, and the insurer prices via frequency--severity, holds reserves, and books IBNR exactly as one would expect. The capped-usage AI contract has the same structure with the dependent variable changed from medical claims, accident severity, or property loss to token-equivalent consumption. The analysis below operates entirely on the seller's side.

\paragraph{Positioning relative to recent work.} \citet{bergemann2026menu} derive the optimal capped menu for LLMs via multidimensional screening, with a risk-neutral buyer and no insurance vocabulary. Their derivation is consistent with the framing above: even if the buyer is risk-neutral and the cap originates as a screening device rather than as risk transfer, the seller's portfolio variance still requires capital. The two perspectives are complementary at the level of mechanism: BBS shows why the cap arises under one principle; this paper shows how the seller manages the resulting exposure under any principle. The same complementarity holds against the broader nonlinear-pricing and three-part-tariff literatures \citep{lambrecht2007uncertainty,bhargava2018optimality}: those works characterize the menu---included allowance, marginal price above the allowance, fixed access fee---that the seller offers. This paper takes any committed menu and treats the resulting aggregate exposure as an economic-capital problem under tail risk and cross-user dependence, which is what those works do not do. A small literature applies actuarial premium principles to cloud SLA outage as an insurance product sold to the cloud service provider \citep{mastroeni2019service,mastroeni2022pricing,bahl2025computation}, modeling downtime as the stochastic variable. The framework here transposes the same machinery to consumption as the stochastic variable, which is the case relevant for capped-usage SaaS pricing. The behavioral pricing literature has identified an ``insurance effect'' in flat-rate tariff choice as a consumer-side bias \citep{lambrechtskiera2006paying}; we operate on the firm side, where the toolkit is reserve calculation and tail-risk capital, not behavioral preference elicitation. An early analogical observation that capped cloud services share structural features with insurance products appears in \citet{weinman2011melons} as a working paper; that essay is conceptual rather than operational and predates the LLM subscription category.

\paragraph{Roadmap.} Section~\ref{sec:structure} formalizes the actuarial structure. Section~\ref{sec:mapping} maps the framework to publicly observable capped-usage SaaS tiers. Section~\ref{sec:modeling} surveys the modeling decisions and open problems a practitioner faces when implementing this in production, ending with a Monte Carlo worked example. Section~\ref{sec:notes} collects practical implementation notes. Section~\ref{sec:conclusion} concludes.

\section{The Actuarial Structure}\label{sec:structure}

We work with a portfolio of $n$ users indexed $i = 1, \ldots, n$, each holding a subscription contract over a reset period $T$ (weekly or monthly in current practice). Each user pays a fixed premium $P_i$ at the start of the period, decoupled from realized consumption. Within the period, user $i$ generates a random count of events $N_i$ with non-negative integer support, where an ``event'' is the contract's unit of billable consumption: a prompt, a session, an API call, or a tool invocation, depending on product. Each event $j$ carries a non-negative severity $S_{ij}$ denominated in the provider's cost units, typically tokens weighted by a model-specific multiplier or directly in dollars at the marginal API rate. The contract imposes a non-fungible cap $K_i$: once cumulative consumption reaches $K_i$, service degrades or halts until $T$ elapses and the cap resets. We treat $K_i$ as exogenous and tier-specific, with downstream consequences left to Section~\ref{sec:modeling}.

The primitive of the model is the per-user aggregate cost over a period:
\begin{equation}\label{eq:aggregate}
C_i \;=\; \min\!\left(K_i, \; \sum_{j=1}^{N_i} S_{ij}\right).
\end{equation}
This is a censored compound random variable. Without the $\min(\cdot, K_i)$ operator, $C_i$ is a compound distribution familiar from non-life insurance: $S_i^{\mathrm{agg}} = \sum_{j=1}^{N_i} S_{ij}$ with $N_i$ from a count family and $\{S_{ij}\}_{j=1}^{N_i}$ i.i.d.\ from a severity family conditional on $N_i$. The cap turns the right tail of $S_i^{\mathrm{agg}}$ into a point mass at $K_i$, which we discuss in Section~\ref{sec:censoring}.

The portfolio aggregate cost is
\begin{equation}\label{eq:portfolio}
L \;=\; \sum_{i=1}^{n} C_i.
\end{equation}
With $\{C_i\}$ independent across users (a working assumption to be relaxed in Section~\ref{sec:catastrophe}), $L$ is a sum of $n$ independent censored compound variables. Two adequacy conditions govern the contract economics. \textbf{Premium adequacy} requires that total premium income covers expected loss with a safety loading $\eta > 0$:
\begin{equation}\label{eq:premium}
\sum_{i=1}^{n} P_i \;\geq\; (1+\eta)\, \mathbb{E}[L].
\end{equation}
\textbf{Reserve adequacy} requires that the provider holds an additional reserve $R$ such that the probability of an aggregate shortfall over the period is bounded by a chosen risk tolerance $\alpha$:
\begin{equation}\label{eq:reserve}
\mathbb{P}\!\left(L \;>\; \textstyle\sum_i P_i + R\right) \;\leq\; \alpha.
\end{equation}
Equivalently, $R = \mathrm{VaR}_{1-\alpha}(L) - \sum_i P_i$ when this quantity is positive. Tail-Value-at-Risk gives a coherent alternative when the tail beyond $\mathrm{VaR}$ is non-negligible, as it typically is for heavy-tailed severity. We return to reserve dimensioning in Section~\ref{sec:reserve}.

These three equations---\eqref{eq:aggregate}, \eqref{eq:premium}, \eqref{eq:reserve}---are the operational identity of an insurance contract restated for a capped-usage subscription. Nothing in the structure depends on legal classification. The remainder of this section specifies the modeling components.

\subsection{Frequency}

The count distribution $F_N$ governs how often a user invokes the service within $T$. The Poisson model $N_i \sim \mathrm{Poisson}(\lambda_i)$ with a per-user rate $\lambda_i$ is the natural starting point but is rarely defensible empirically: usage data on capped-usage subscriptions exhibits overdispersion (variance materially exceeding mean), inflated zeros (a non-trivial subpopulation does not engage in a given period), and bursty within-period patterns. Three remedies cover the typical empirical pattern: a Negative Binomial $N_i \sim \mathrm{NB}(r, p_i)$ for overdispersion, a Zero-Inflated Negative Binomial for excess zeros, and a Hurdle model when the zero-generating process is structurally distinct from the positive-count process. Latent class or hierarchical Bayesian specifications \citep{klugman2012loss} accommodate unobserved user segments without requiring an external segmentation variable. The choice between these is empirical and tier-specific: a Max~20x cohort has different overdispersion structure than a Pro cohort.

\subsection{Severity}

The severity distribution $F_S$ governs the cost of a single event. Two families are natural in this domain. \textbf{Gamma severity}, $S_{ij} \sim \mathrm{Gamma}(\alpha, \beta_i)$, is analytically convenient: under a Poisson frequency, the compound $S_i^{\mathrm{agg}}$ admits a Tweedie representation with closed-form moments, and the conjugate Poisson--Gamma family enables hierarchical credibility estimation in the B\"uhlmann--Straub tradition. \textbf{LogNormal severity}, $\log S_{ij} \sim \mathcal{N}(\mu_i, \sigma^2)$, accommodates the right skew and multiplicative scaling typical of token consumption per event, particularly when long chain-of-thought outputs or multi-turn tool invocations dominate the right tail. For catastrophe modeling (Section~\ref{sec:catastrophe}), a Generalized Pareto tail above a threshold $u$ is standard. The choice is informed by goodness-of-fit on observed (uncensored) events.

LLM-specific severity has a structural advantage rare in classical applications: it decomposes additively into observable components. For a prompt $j$ to user $i$:
\begin{equation}\label{eq:llmseverity}
S_{ij} \;=\; c_{\text{in}} \cdot \text{tokens}_{\text{in},ij} + c_{\text{out}}(m_{ij}) \cdot \text{tokens}_{\text{out},ij} + \sum_{k} c_{\text{tool},k} \cdot \mathbb{1}[\text{tool}_k \in j],
\end{equation}
where $c_{\text{in}}, c_{\text{out}}$ are per-token costs (with $c_{\text{out}}$ depending on the model class $m_{ij}$, e.g., a 5x multiplier for Opus relative to Sonnet), and tool-call costs $c_{\text{tool},k}$ aggregate compute and dependency costs of agentic invocations. Health-insurance severity has no analogue of~\eqref{eq:llmseverity}: one cannot decompose a medical claim into $c_{\text{in}} \cdot x_{\text{in}}$ ex ante. This decomposition gives capped-usage AI providers a modeling lever that the actuarial literature on health and auto lines does not possess, and we revisit it in Section~\ref{sec:modeling}.

\subsection{Compound distributions}

The unconditional distribution of $S_i^{\mathrm{agg}} = \sum_{j=1}^{N_i} S_{ij}$ is the convolution of the severity distribution under random replication by $N_i$. Two cases are tractable:

\begin{itemize}
\item \textbf{Poisson--Gamma.} When $N_i \sim \mathrm{Poisson}(\lambda_i)$ and $S_{ij} \sim \mathrm{Gamma}(\alpha, \beta_i)$, $S_i^{\mathrm{agg}}$ has a Tweedie distribution with parameters indexed by $(\lambda_i, \alpha, \beta_i)$. Mean and variance are available in closed form; the density requires numerical evaluation but is well-supported in modern GLM implementations (\texttt{tweedie} in R, \texttt{statsmodels} GLM family in Python).
\item \textbf{NB--LogNormal.} When $N_i \sim \mathrm{NB}(r, p_i)$ and $\log S_{ij} \sim \mathcal{N}(\mu_i, \sigma^2)$, the compound has no closed form and is evaluated by Monte Carlo. This is the workhorse case in Section~\ref{sec:worked} and in the companion \texttt{sims/} repository.
\end{itemize}

Other combinations (NB--Gamma, Poisson--LogNormal, Hurdle--Pareto) are accessible by Monte Carlo without further conceptual machinery.

\subsection{Censoring at the cap}\label{sec:censoring}

The cap converts $S_i^{\mathrm{agg}}$ into the censored variable $C_i$ defined in~\eqref{eq:aggregate}. Observed data on $C_i$ is right-censored at $K_i$ for the subpopulation that hits the cap within $T$; this subpopulation is precisely the heavy-usage tail that drives reserve adequacy. Two conceptually distinct things are happening under censoring, depending on the contract regime. In a true policy-limit insurance contract (and in soft-cap SaaS regimes such as ChatGPT model-class restriction), the underlying severity event still occurs above $K_i$ and the insured absorbs the dollar excess. In hard-cap SaaS (Claude Code post-August 2025), demand above $K_i$ is \emph{latent}: the service halts, the user's tail consumption is not served, and what the user absorbs is unavailability and postponement rather than dollar cost. The actuarial machinery for fitting the latent distribution is the same in both cases (Tobit-style MLE with explicit censored-data likelihood contributions, or estimation on the event-level data $\{S_{ij}\}$ restricted to events occurring before the user reaches $K_i$ with a survival-style correction for the truncation), but the interpretation of the fitted tail differs: in the soft-cap and policy-limit cases the tail describes realized consumption that someone pays for; in the hard-cap case it describes the demand the user \emph{would} have realized absent the cap. Both are operationally relevant for the seller (the first determines exposure, the second determines pricing-and-retention behavior), and both are textbook \citep{wuthrich2023nonlife,frees2010regression}. We treat the bias quantitatively in Section~\ref{sec:severity-bias}.

\subsection{Portfolio loss and reserve}

Equations~\eqref{eq:aggregate}--\eqref{eq:reserve} close the structure: given parameterized frequency and severity models, the portfolio loss $L$ can be simulated, and the reserve $R$ satisfying~\eqref{eq:reserve} computed. Three observations matter for downstream sections.

First, $\mathbb{E}[L]$ scales linearly in $n$ while $\mathrm{Var}[L]$ scales linearly in $n$ under independence, so the coefficient of variation of $L$ scales as $n^{-1/2}$: large portfolios benefit from the law of large numbers, and per-user reserve $R/n$ falls with portfolio size, as in any insurance line.

Second, the cap $K_i$ caps the contribution of each user to $L$ at $K_i$, which bounds the tail of $L$ from above by $\sum_i K_i$. Reserve adequacy can never require more than this bound, but the bound is loose; the operative quantity is the realized aggregate, not the contractual ceiling.

Third, the cap reset schedule $T$ creates structural dependence across periods: a user who hits the cap in period $t$ has a behavioral response in period $t+1$ that the within-period model does not capture. This is the behavioral component formalized in Section~\ref{sec:behavioral}.

\subsection{Contract regimes: hard cap, soft degradation, and included-plus-overage}\label{sec:regimes}

Equation~\eqref{eq:aggregate} writes per-user cost as $C_i = \min(K_i, S_i^{\mathrm{agg}})$. This is the right object for the hard-cap regime exemplified by Claude Code Max post-August 2025: service halts at $K_i$, and the user-side cost in dollars is bounded by $K_i$ regardless of what the user would have consumed in the absence of the cap. Two other regimes are operationally common and require small adjustments to the same framework; we treat all three uniformly as compound-loss models with regime-specific truncation or top-up.

\textbf{Hard cap.} Service halts at $K_i$. Seller exposure per user is bounded above by $K_i$. The full~\eqref{eq:aggregate}--\eqref{eq:reserve} apply directly with $C_i = \min(K_i, S_i^{\mathrm{agg}})$.

\textbf{Soft degradation.} Service does not halt at $K_i$; instead the user is downgraded to a cheaper model class, lower priority queue, or rate-limited tier. Seller exposure is $C_i = \min(K_i, S_i^{\mathrm{agg}}) + \rho \cdot \max(0, S_i^{\mathrm{agg}} - K_i)$ with $\rho \in (0, 1)$ the cost ratio of the degraded service to the primary service. Reserve adequacy depends on both $\rho$ and the realized post-degradation severity.

\textbf{Included-plus-overage.} The cap is a price kink, not a service halt. Above $K_i$, the user pays a posted marginal price $r_i$ per unit of consumption; the seller incurs cost on the realized usage and books overage revenue. The relevant accounting object is net loss
\begin{equation}\label{eq:overage}
L_i^{\text{net}} \;=\; \mathrm{cost}_i - P_i - r_i \cdot \max(0, S_i^{\mathrm{agg}} - K_i),
\end{equation}
where $\mathrm{cost}_i$ is the seller's realized cost (typically $\approx S_i^{\mathrm{agg}}$ at marginal cost, possibly with a markup factor below 1 reflecting the gap between posted retail price and provider marginal cost). Reserve adequacy in this regime is the requirement that the portfolio's net loss has a controlled tail; the operative quantity shifts from cap dimensioning to overage-pricing adequacy ($r_i$ chosen so that $\mathbb{E}[L_i^{\text{net}}] \leq 0$ for the high-usage subpopulation). Vercel Pro, Cloudflare Workers, Supabase, and GitHub Copilot operate in this regime.

The three regimes share \emph{compound severity-frequency structure}, \emph{censoring or kinked-price effects at $K_i$}, and \emph{aggregate tail risk that determines reserve}. The fitting machinery of Section~\ref{sec:severity-bias} and the reserve-dimensioning logic of Section~\ref{sec:reserve} apply across all three with the per-user object adapted to the regime. The remainder of this paper develops examples primarily in the hard-cap and included-plus-overage cases.

\section{Mapping to Real Products}\label{sec:mapping}

The structure of Section~\ref{sec:structure} is not a stylization. The currently dominant capped-usage SaaS tiers instantiate~\eqref{eq:aggregate}--\eqref{eq:reserve} with publicly observable parameters. Table~\ref{tab:products} maps the most-discussed products to the primitives of the model, using provider pricing pages and public documentation accessed in May 2026.

\begin{table}[t]
\centering
\small
\setlength{\tabcolsep}{5pt}
\begin{tabular}{@{}p{2.9cm}p{1.3cm}p{2.2cm}p{3.6cm}p{2.4cm}p{1.3cm}@{}}
\toprule
\textbf{Product} & \textbf{Premium $P$} & \textbf{Event unit} & \textbf{Severity $S_{ij}$} & \textbf{Cap $K$} & \textbf{Reset $T$}\\
\midrule
\multicolumn{6}{l}{\emph{LLM services}} \\
Claude Code Pro & \$20/mo & Prompt/session & Tokens; $c_{\mathrm{out}}$ depends on model (Opus $\approx 5\times$ Sonnet) & Weekly token budget (Aug 2025+) & Weekly \\
Claude Code Max & \$100/mo & Prompt/session & Same as Pro & $\approx 5\times$ Pro budget & Weekly \\
Claude Code Max 20x & \$200/mo & Prompt/session & Same as Pro & $\approx 20\times$ Pro budget & Weekly \\
ChatGPT Plus & \$20/mo & Message & Tokens, screened by model class (o-tier restricted) & Rolling soft caps per model & Rolling \\
ChatGPT Pro & \$200/mo & Message & Tokens, full model menu incl.\ o1-pro & Rolling soft caps, higher & Rolling \\
GitHub Copilot & \$10--39/mo & Completion / chat turn & Tokens, overage charged at \$/M & Floor included, overage billed & Monthly \\
\addlinespace[0.4ex]
\multicolumn{6}{l}{\emph{Cloud platforms}} \\
Vercel Pro & \$20/mo & Deploy / fn-call & Bandwidth bytes, build-min, function GB-sec & 1\,TB BW + 6k build-min & Monthly \\
Cloudflare Workers & \$5/mo & Worker request & Request count, CPU time & 10M req incl., overage \$/M & Monthly \\
Supabase Pro & \$25/mo & DB query / row / MAU & DB ops, storage GB, MAU & 8\,GB DB + 100\,GB storage + 100k MAU & Monthly \\
\bottomrule
\end{tabular}
\caption{Public capped-usage SaaS subscriptions, May 2026, mapped to the actuarial primitives $(P, N_i, S_{ij}, K, T)$ of Section~\ref{sec:structure}. Sources: provider pricing pages and public documentation accessed May 2026 (\href{https://www.anthropic.com/pricing}{anthropic.com/pricing}, \href{https://openai.com/api/pricing/}{openai.com/api/pricing}, \href{https://github.com/features/copilot/plans}{github.com/features/copilot/plans}, \href{https://vercel.com/pricing}{vercel.com/pricing}, \href{https://developers.cloudflare.com/workers/platform/pricing/}{Cloudflare Workers pricing docs}, \href{https://supabase.com/pricing}{supabase.com/pricing}). Parameter values change quarterly; the structural mapping does not.}
\label{tab:products}
\end{table}

The variation across products is parametric, not structural. Claude Code and ChatGPT differ on reset cadence (weekly versus rolling) and on whether the cap is hard (service halts) or soft (degraded model class). GitHub Copilot, Vercel, Cloudflare Workers, and Supabase invert the constraint relative to Claude Code: instead of a hard service halt at $K$, they set a fixed premium guaranteeing baseline consumption with overage billed at a posted rate above $K$. \citet{bergemann2026menu} characterize the Maximum-Spend and Minimum-Spend mechanisms as the two canonical menu items in this design space. Both fit~\eqref{eq:aggregate}--\eqref{eq:reserve}: in the hard-cap case, $C_i = \min(K, S_i^{\mathrm{agg}})$ truncates exposure on the seller's side; in the overage case, the seller's exposure remains $S_i^{\mathrm{agg}}$ but premium income tracks realized cost above $K$, shifting the operative quantity from reserve adequacy to overage-pricing adequacy.

\subsection{A non-LLM example: Vercel Pro}\label{sec:vercel}

Vercel Pro is a deployment platform for web applications priced at \$20 per developer per month with the cap structure stated in Table~\ref{tab:products}. We use it as the second worked example because it exhibits the same actuarial structure as a Claude Code tier with three differences that illuminate the framework's scope.

\paragraph{The event is not a prompt.} The natural event unit for a Vercel subscriber is a deployment combined with the public traffic served against that deployment over the period: the build cost (compute minutes), the function invocation cost (GB-seconds of serverless compute), and the bandwidth cost (gigabytes egressed). Severity $S_{ij}$ is denominated in dollar-equivalent at posted overage rates: \$0.15 per GB bandwidth, \$0.60 per million function invocations, \$0.128 per CPU-hour, and roughly \$0.014 per minute of build execution on standard machines as of May 2026.\footnote{Source: \href{https://vercel.com/pricing}{vercel.com/pricing} accessed May 2026.} Frequency $N_i$ is the number of deployment-plus-traffic events in a month, which for a solo developer hosting a personal site is a small number and for an e-commerce shop with a viral release is on the order of hundreds.

\paragraph{Severity does not decompose as in \eqref{eq:llmseverity}.} The LLM-specific decomposition into input tokens, output tokens, model multiplier, and tool calls has no Vercel analogue. Instead, Vercel severity is the linear combination $c_{\mathrm{bw}} \cdot \mathrm{GB}_{ij} + c_{\mathrm{fn}} \cdot \mathrm{GBh}_{ij} + c_{\mathrm{bld}} \cdot \mathrm{min}_{ij}$, with the components determined by application behavior rather than by model class. The implication is that \eqref{eq:llmseverity} is LLM-specific while \eqref{eq:aggregate}--\eqref{eq:reserve} are not, which is the modular structure of the framework we claim throughout.

\paragraph{The cap is multi-axis.} Vercel Pro caps three quantities simultaneously (bandwidth, function compute, build minutes), each with its own per-axis overage rate. Formally, $C_i = \sum_{a \in \mathrm{axes}} c_a \cdot \max(0, X_{ia} - K_{ia}) + \mathrm{(included)}$, which is a vector-valued retention $K_i \in \mathbb{R}^3_+$ rather than a scalar. The same framework applies, with $K_i$ a vector and the censoring of Section~\ref{sec:censoring} applied per axis. This is closer to the multi-line reinsurance contracts than to the single-line health policies invoked in Section~\ref{sec:reserve}, and it is operationally what a typical SaaS contract looks like in practice. The expectation $\mathbb{E}[L]$ separates additively across axes only if axis-level consumption is independent, which is rarely true; bandwidth and function calls correlate by application architecture. The textbook scalar Tobit MLE of Section~\ref{sec:severity-bias} therefore fails to recover the joint severity distribution in the multi-axis case, and the practitioner needs a multivariate-Tobit or copula-based censored-likelihood approach. Multivariate Tobit identification is weak in small samples and computationally heavier than the scalar case, so this is a genuine open problem rather than a textbook application.

\paragraph{Heavy-user illustration.} As an order-of-magnitude calibration consistent with public Vercel pricing, take a Vercel Pro subscriber population of $n = 10{,}000$ developers, premium \$20/seat/month each, expected bandwidth consumption per developer of 100\,GB/month with a LogNormal tail. The published bandwidth allowance at the Pro tier is 1\,TB included with \$0.15/GB overage. A single subscriber with a viral release week consuming 5\,TB pays a marginal \$0.15 $\times$ (5{,}000 $-$ 1{,}000) = \$600 in overage on top of the \$20 premium, while the median subscriber consuming 50\,GB pays the premium only. The seller's exposure is \emph{not} bounded at $K = 1\,\mathrm{TB}$ because the overage is billed; the operative actuarial question for Vercel is overage-pricing adequacy, not reserve adequacy at the cap.

\paragraph{Monte Carlo under the overage regime.} Table~\ref{tab:vercel} reports a Monte Carlo of the Vercel-like overage scenario described above, $n = 10{,}000$, premium \$20/period, cap allowance \$1{,}000 (at retail-equivalent value), posted overage rate $r = 0.15$ (fraction of retail). We compare two cohorts within the same framework: a \emph{light} cohort (median developer, mean consumption \$45/user, below the cap) and a \emph{heavy} cohort (e-commerce or viral-release subscriber, mean consumption \$1{,}114/user, above the cap). Because the seller's actual marginal cost is private and lower than the retail rate, we sweep the cost-to-retail ratio $\kappa \in \{0.25, 0.50, 1.00\}$ within each cohort.

\begin{table}[t]
\centering
\small
\setlength{\tabcolsep}{5pt}
\begin{tabular}{@{}llrrrrr@{}}
\toprule
\textbf{Cohort} & $\kappa$ & $\mathbb{E}[S^{\mathrm{agg}}]$ & $\mathbb{E}[\mathrm{cost}]$ & Premium & $\mathbb{E}[\mathrm{overage~rev}]$ & $\mathbb{E}[L^{\mathrm{net}}]$ \\
\midrule
Light & 0.25 & \$451k & \$113k & \$200k & \$0 & $-$\$87k (surplus) \\
Light & 0.50 & \$451k & \$226k & \$200k & \$0 & $+$\$26k (slight loss) \\
Light & 1.00 & \$451k & \$451k & \$200k & \$0 & $+$\$251k (deep loss) \\
Heavy & 0.25 & \$11.13M & \$2.78M & \$200k & \$658k & $+$\$1.92M (loss) \\
Heavy & 0.50 & \$11.14M & \$5.57M & \$200k & \$659k & $+$\$4.71M (deep loss) \\
Heavy & 1.00 & \$11.13M & \$11.13M & \$200k & \$658k & $+$\$10.28M (deep loss) \\
\bottomrule
\end{tabular}
\caption{Vercel-like overage scenario, Monte Carlo over $2{,}000$ replications, two cohorts at fixed overage rate $r = 0.15$. $\kappa$ is the cost-to-retail ratio (seller's actual marginal cost divided by retail price). In the light cohort, mean consumption (\$45/user) is far below the cap allowance (\$1{,}000), so virtually no user triggers overage and the contract behaves operationally as a flat-rate subscription. In the heavy cohort, mean consumption (\$1{,}114/user) exceeds the cap on average, triggering substantial overage revenue (\$658k per period), but the posted overage rate is insufficient to cover marginal cost even at $\kappa = 0.25$. Solvency in expectation depends on cohort composition and $\kappa$ jointly. Generated by \texttt{sims/studies/study\_vercel.py}.}
\label{tab:vercel}
\end{table}

Three readings of Table~\ref{tab:vercel} matter. First, the overage mechanism is empirically inactive for the light cohort (overage revenue rounds to zero) and active but insufficient for the heavy cohort (overage revenue \$658k against a cost gap of several million USD). The contract behaves bimodally across cohorts: a flat-rate subscription for one and a loss-making per-unit service for the other. Second, in the heavy cohort the posted overage rate $r = 0.15$ does not match the seller's economics even at $\kappa = 0.25$; the operational implication is that overage pricing must be calibrated to the cohort-conditional severity tail, not to the population-average consumption. Third, solvency in expectation depends jointly on cohort composition and $\kappa$. This is the operative tension in cloud-platform pricing that does not appear in the hard-cap regime, where seller exposure is bounded by $K$ regardless of $\kappa$. The actuarial framework applies in both regimes, but the diagnostic statistics differ: $\mathrm{TVaR}(L)$ for hard cap, $\mathbb{E}[L^{\mathrm{net}}]$ jointly across cohort and $\kappa$ for overage.

The calibration code is in \texttt{sims/src/saas\_actuaria/calibration.py} under \texttt{vercel\_scenario()} and the Monte Carlo is \texttt{sims/studies/study\_vercel.py}. The retail-cost decomposition uses $\kappa$ as a free parameter rather than an estimate, for reasons explained in Section~\ref{sec:cost-vs-retail}.

\subsection{Retail price, marginal cost, and the \texorpdfstring{$\kappa$}{kappa} ratio}\label{sec:cost-vs-retail}

The Vercel calibration in Table~\ref{tab:vercel} surfaces a measurement issue that recurs across capped-usage SaaS pricing analysis. Observed quantities for an outside analyst are typically denominated at the seller's posted retail rate (\$/GB bandwidth, \$/1M tokens, \$/check-in). The seller's actual marginal cost is private and lower, often substantially so for AI services where inference compute has been the subject of years of optimization. We denote the ratio of marginal cost to posted retail rate as $\kappa \in (0, 1]$.

Two practical consequences follow. First, exposure figures reported at retail (Table~\ref{tab:reserve} and Table~\ref{tab:policy}) are upper bounds on the seller's economic exposure. Where the qualitative direction of an argument matters more than the magnitude---for instance, that no-cap subscription tiers are infeasible at any plausible $\kappa$ under sufficiently heavy consumption---the framework conclusions are robust to $\kappa$. Where magnitudes matter directly---specifically the headline ``loss ratio'' in Table~\ref{tab:policy} P2---the figure should be read as an upper bound and not as a forecast of realized economic loss. Anyone with access to private cost data should refit the worked examples under $\kappa < 1$.

Second, in the overage regime (Table~\ref{tab:vercel}) the operative actuarial diagnostic is gross-margin solvency in expectation rather than tail-aware reserve adequacy, and the diagnostic is exactly the value of $\kappa$ relative to the posted overage rate. Reserve calculations on $L^{\mathrm{net}}$ remain well-defined, but the headline question shifts from ``does the cap absorb the tail'' to ``does retail-rate revenue cover marginal cost at portfolio scale''. The actuarial framework applies; what changes is which equation is the binding one.

A parallel observation runs in the opposite direction: insurance practice itself is increasingly adopting usage-based pricing structures, as documented by \citet{holzapfel2024mitigating} for auto and \citet{zeller2021comprehensive} for cyber. The cross-traffic between SaaS pricing and insurance pricing is bidirectional; this paper traffics in the SaaS-as-insurance direction, but the insurance-as-usage-based direction is the converse.

Endogeneity of the consumption distribution to the seller deserves a related caveat. In classical non-life insurance, the loss distribution $F_N, F_S$ is taken as exogenous conditional on the contract, and the seller exercises control through acceptance criteria (underwriting), contract terms (exclusions, deductibles, limits), and pricing (ratemaking). The same seller-side levers apply in capped-usage SaaS. The difference is that the SaaS seller additionally exercises product-design levers (latency, agentic defaults, UX, model-release cadence) that shift the conditional distribution \emph{post-acceptance}, where in classical insurance the claim event is taken as exogenous conditional on the contract. The actuarial parallel holds at the level of compound loss structure and at the level of seller-side controls already accommodated by insurance practice; the additional product-design margin is a genuine difference that the practitioner should hold in mind when refitting the distribution after a product change. It does not undermine the framework; it makes the refit cadence and the model-cost-shock discussion in Section~\ref{sec:reserve} more important.

Two further product families exhibit the same structure and motivate the framework's full scope.

\paragraph{Corporate benefit platforms with ``unlimited'' access.} Platforms offering fitness, education, telemedicine, or wellness services to enterprise buyers on a flat-fee-per-employee basis with capped or rate-limited usage operate the same contract. The fitness benefit category (Wellhub, formerly Gympass; ClassPass enterprise) is the most-studied. The premium is a per-employee monthly fee, the event is a class booking or facility check-in, the severity is a per-event payment to the partner gym, and the cap is enforced as a frequency limit or a tier-based credit allowance. The dependent variable is check-ins, not tokens or bandwidth, but $C_i$, $L$, and reserve dimensioning are unchanged.

\paragraph{SaaS products with embedded liability transfer.} Identity verification, fraud detection, and chargeback-guarantee products charge a fixed fee per validation but assume a contractual exposure for false negatives that exceeds the fee by orders of magnitude. The premium is per-API-call, the event is a fraudulent transaction that the validation failed to block, and the severity is the dollar value of the resulting chargeback or breach. The cap is implicit in the contractual exposure cap or in the firm's solvency. Reserve adequacy in this case is no longer optional in any operational sense, even when the regulatory regime does not classify the product as insurance.

The point is scope, not novelty. The argument of this paper is that the same machinery---premium adequacy, censored compound distributions, reserve and overage adequacy via tail risk---applies to all four contract families above. The LLM subscription case in Section~\ref{sec:worked} and the cloud platform case in Section~\ref{sec:vercel} are the two most public and most rapidly evolving instances; the corporate benefits and liability transfer cases are older instances of a category that has always been actuarial.

\section{Modeling Decisions and Open Problems}\label{sec:modeling}

The structure of Section~\ref{sec:structure} leaves seven modeling choices open. We treat them as practical decisions rather than theoretical claims, flag the actuarial machinery available for each, and identify open problems specific to the AI subscription setting.

\subsection{Frequency distribution choice}

Observed event counts on AI subscriptions exhibit two simultaneous features that rule out the Poisson default: variance exceeding mean by a factor of 3--20 in publicly reported usage telemetry, and bimodality between a quiet majority and a heavy-usage minority. Negative Binomial accommodates the first; Zero-Inflated Negative Binomial or Hurdle models accommodate the joint pattern. When power-user identification is the goal rather than a nuisance, the Hurdle decomposition is operationally cleaner: the zero-process is a binary engagement model (logistic regression on user features) and the positive-count process is a truncated NB on engaged users. Latent class extensions \citep[ch.~6]{klugman2012loss} and hierarchical Bayesian specifications recover the segmentation without requiring exogenous user labels, which matters because tier choice itself is endogenous (see Section~\ref{sec:adverse}).

An open problem specific to AI is that the unit of an ``event'' is not pinned by the contract. A single Claude Code session can span hundreds of tool calls and dozens of model invocations; under one definition $N_i$ counts sessions, under another it counts tool calls, and the implied severity distribution differs accordingly. Standard actuarial practice in health and auto lines treats the claim as the unit; the analogue here is a design choice, not data.

\subsection{Severity treatment under censoring}\label{sec:severity-bias}

Section~\ref{sec:censoring} introduced the censoring effect of the cap. Quantitatively, naive maximum likelihood fit of a severity distribution to observed $\{S_{ij}\}$ values truncated by the contract's effective cap underestimates both the location and the scale of the underlying distribution, with the bias monotone in the fraction of users hitting $K_i$ within $T$. For LogNormal severity with true parameters $\mu = 2.6, \sigma = 1.3$ and a sample of $n = 50{,}000$ severities, a Monte Carlo study yields a downward bias of $-5.5\%$ in $\hat\mu$ and $-10.2\%$ in $\hat\sigma$ at 5\% censoring, growing to $-17.7\%$ and $-24.0\%$ at 20\% censoring, and to $-32.1\%$ and $-35.4\%$ at 40\% censoring (\texttt{sims/studies/study\_censoring\_bias.py}). A Tobit-style censored MLE recovers parameters to within 1\% in all three regimes. The implication for reserve adequacy is that naive estimation understates the true tail by a multiplicative factor that grows with $\sigma$ and with the censoring fraction. The standard remedies are Tobit-style MLE with explicit censored-data likelihood contributions, or estimation restricted to events occurring before user $i$ reaches $K_i$ with an appropriate truncation correction. Both are textbook \citep{wuthrich2023nonlife,frees2010regression} and available in standard libraries: \texttt{actuar} in R; in Python, the survival-analysis package \texttt{lifelines} covers Tobit-style truncation directly, with custom MLE on \texttt{scipy.optimize} as a manual alternative.

The open problem is that the censoring point $K_i$ is not always observable to the modeler in the form used by the contract. Anthropic's weekly cap, for example, is denominated in a proprietary ``usage units'' function of tokens and model class, not in tokens directly; modeling severity in dollars requires reconstructing the cap function from public pricing announcements, which is feasible but introduces measurement noise.

\subsection{Heterogeneity and adverse selection}\label{sec:adverse}

A user choosing the Max~20x tier at \$200 per month is not a random draw from the population of potential subscribers. Tier choice is endogenous to expected consumption, and the conditional severity distribution within a tier reflects both this self-selection and the behavioral response to the higher cap. \citet{einavfinkelstein2011selection} is the canonical survey of selection in insurance markets, the earlier empirical strategy in \citet{cutler1998paying} is the foundational competition-vs-selection treatment, and the modern empirical identification strategy is the Einav--Finkelstein cost curve \citep{einav2010estimating}, which uses price variation to identify the marginal cost of insuring an additional unit of risk. Two effects compound the identification problem in capped-usage SaaS specifically. First, adverse selection: heavy users self-select into higher tiers. Second, plan inertia: \citet{handel2013inertia} documents that consumers underrespond to plan-design changes, holding on to suboptimal tiers and dampening cross-sectional price elasticity in the short run; \citet{einav2025selling} extends the same observation to general subscription products including digital services. Both effects are first-order for capped-usage SaaS, where tier choice is sticky and tier-switching events are sparse.

\paragraph{What heterogeneity does to the tail.} The default Monte Carlo of Section~\ref{sec:worked} uses a single NB-LogNormal whose overdispersion implicitly absorbs unobserved heterogeneity (the NB is itself a Gamma-mixed Poisson, equivalent to a model in which each user has a private rate). This is the right specification when the modeling team has no segmentation variable. When segmentation \emph{is} observable---a tier label, an enterprise vs.\ self-serve flag, a power-user behavioral signature---the appropriate model is an explicit mixture. We illustrate the difference in Table~\ref{tab:mixed} with three mixed-population scenarios calibrated so that the portfolio expected loss matches the homogeneous default (\$30/user, \$300k aggregate at $n = 10{,}000$).

\begin{table}[t]
\centering
\small
\begin{tabular}{@{}lrrrrrr@{}}
\toprule
\textbf{Scenario} & $\pi_{\mathrm{power}}$ & $\mathbb{E}[L]$ & $\mathrm{VaR}_{0.99}$ & $\mathrm{TVaR}_{0.99}$ & \textbf{Cap-hit power} & \textbf{Cap-hit light} \\
\midrule
H.\ Homogeneous (default) & --- & \$300k & \$309k & \$310k & --- & --- \\
M1.\ 90\% light / 10\% power & 10\% & \$299k & \$318k & \$322k & 0.5\% & 0.0\% \\
M2.\ 80\% light / 20\% power & 20\% & \$300k & \$314k & \$316k & 0.0\% & 0.0\% \\
M3.\ 95\% light / 5\% power  & 5\%  & \$297k & \$318k & \$320k & 2.4\% & 0.0\% \\
\bottomrule
\end{tabular}
\caption{Mixed-population portfolio statistics at matched expected loss. The light segment has $\mathbb{E}[N] = 5$, $\mathbb{E}[S] \in \{2, 3\}$, $\sigma = 0.8$; the power segment has $\mathbb{E}[N] \in \{22, 30, 45\}$, $\mathbb{E}[S] \in \{5, 7\}$, $\sigma = 1.5$, with the mixture weights chosen so that $(1-\pi_{\mathrm{power}}) \mathbb{E}_L + \pi_{\mathrm{power}} \mathbb{E}_P = 30$. Cap-hit columns report the within-segment fraction of users whose realized period cost equals the cap. Generated by \texttt{sims/studies/study\_mixed\_population.py}.}
\label{tab:mixed}
\end{table}

Three observations follow. First, explicit heterogeneity raises $\mathrm{TVaR}_{0.99}$ by only 2\% to 4\% relative to homogeneous, at identical expected loss---a materially smaller effect than the 20--50\% range typical in classical actuarial lines (auto, health). This is not a failure of the framework; it is the expected consequence of policy-limit design. The hard cap $K_i$ binds before per-user severity reaches the long tail of the power segment, truncating individual exposure before it aggregates to the portfolio. \emph{Adverse selection is a first-order modeling concern but a second-order reserve concern in the hard-cap regime, precisely because the cap is doing exactly the work it is designed to do.} The implication runs in both directions: a practitioner facing a tier with documented heterogeneity should not expect adverse selection alone to drive material reserve adjustment under a tight cap, and a practitioner facing a tier without a tight cap (the overage regime of Section~\ref{sec:vercel}) should expect adverse selection to load directly into the overage-pricing problem instead.

Second, the cap binds almost exclusively on the power segment. Even at the most aggressive mixture (M3, 5\% power users) the within-light-segment cap-hit rate is effectively zero; the within-power-segment rate is 2.4\%. This is the operational signature of adverse selection at the individual level: a small population concentrates the cap-hit exposure, which is what the cap is designed to absorb. Third, the tail is not monotone in $\pi_{\mathrm{power}}$. M3 (5\% concentrated) and M1 (10\% diffuse) give comparable $\mathrm{TVaR}_{0.99}$, because concentration of consumption among a smaller group offsets the smaller weight in the mixture. The pricing implication is that a tier whose customer base contains 5\% of extreme power users and a tier with 10\% moderate power users may require similar reserves; the relevant statistic is the conditional severity distribution among the heavy segment, not its weight alone.

\paragraph{Identification in practice.} The mixture parameters $(\pi_{\mathrm{power}}, F_{\mathrm{light}}, F_{\mathrm{power}})$ are not observed directly. Three identification strategies are available. (i) Propensity-stratified modeling: estimate $F$ within tier-by-cohort cells defined by pre-switch behavior. (ii) Latent class extensions of \citet[ch.~6]{klugman2012loss}: estimate the mixture jointly with the segmentation, without external labels. (iii) Tier-switching events as instruments: when Anthropic introduced weekly limits in August 2025, users above the new cap and users below it form quasi-treatment and quasi-control groups, but their pre-period distributions differ by construction. Naive difference-in-differences violates parallel-trends; valid identification requires either explicitly modeling differential pre-trends (e.g., synthetic control across user-week panels) or a regression-discontinuity design at the cap threshold. The closest empirical work in capped-usage AI \citep{demirer2025emerging} operates at the model level rather than the user level and does not address adverse selection directly.

\subsection{Reserve dimensioning}\label{sec:reserve}

Given a fitted frequency--severity model and an empirical or simulated portfolio aggregate $L$, computing the reserve $R$ satisfying~\eqref{eq:reserve} is a straightforward Monte Carlo exercise. The relevant choices are the risk tolerance $\alpha$ (1\% is conventional, 0.1\% appropriate for catastrophic exposures, see Section~\ref{sec:catastrophe}) and the choice between VaR and TVaR. TVaR is coherent and gives a smoother sensitivity to tail-fitting choices, which matters because the tail beyond $\mathrm{VaR}$ is exactly where modeling uncertainty is highest.

We use ``reserve'' throughout in the operational sense of economic capital or risk buffer held against the aggregate consumption tail---margin retained on the balance sheet, capacity provisioned ahead of expected load, or a working-capital buffer against a bad month---not in the statutory or accounting sense in which an insurer's loss reserves are recognized under specific regulatory regimes. SaaS providers typically do not segregate capital under an insurance framework, but the quantity that makes the contract solvent in a bad period is the same.

For LLM subscriptions specifically, three additional considerations arise. First, model-cost shocks (a new training run, a hardware shift, an inference-stack optimization) introduce non-stationarity in $S_{ij}$ that requires reserve adjustment between fitting epochs; this is the analogue of social inflation in casualty insurance. Second, the cap acts as a policy limit on the seller's exposure: once $K_i$ is hit, the user absorbs further consumption rather than the seller, exactly as in dental annual benefit maximums, auto liability limits, or reinsurance excess-of-loss treaties. The seller's exposure is bounded above by the unceded portion $\min(K_i, S_i^{\mathrm{agg}})$, and reserve calculations operate on this bounded exposure. Third, the reset period $T$ creates a serial dependence in $L_t$ across periods through the behavioral channel of Section~\ref{sec:behavioral}; standard reserve calculations assume independence across periods, and the behavioral correction modifies this.

\paragraph{Operational reserve update.} A reserve $R_t$ for period $t$ is set from an estimated distribution of $L_t$ fitted on previous periods. Three update rules are standard. \emph{Rolling-window MLE}: refit frequency and severity on the most recent $W$ periods of data; choose $W$ to trade off bias (long windows include stale model-cost regimes) against variance (short windows leave parameter estimates noisy). \emph{Exponential smoothing}: weight historical periods by $\rho^{t-s}$ for some $\rho \in (0, 1)$, which down-weights distant history without a hard cutoff. \emph{Hierarchical Bayesian update}: maintain a posterior over $(F_N, F_S)$ that is sequentially updated each period, with a prior chosen to encode the structural-shock concern; under conjugate priors (Gamma-Poisson, inverse-Gamma-LogNormal) this reduces to weighted-average updates of sufficient statistics. The practitioner-relevant choice is between the second and the third: exponential smoothing is operationally simpler and gives smoother reserves; Bayesian update gives a coherent uncertainty quantification on the reserve itself. For a modeling team accustomed to fitting GLMs in production, exponential smoothing on a Tweedie GLM is the lower-friction starting point.

\paragraph{IBNR-analog at period close.} Standard IBNR (incurred-but-not-reported) reserving handles the gap between when a claim is incurred and when it is reported. The capped-usage analog is the gap between when a user-initiated event consumes service and when that consumption is reflected in the billing system. For LLM subscriptions specifically, three IBNR-like sources matter: in-flight agentic sessions at period close (a long-running tool chain spanning the reset boundary), retried failed calls that consume tokens without immediate billing attribution, and asynchronous tool invocations whose cost lands after the user's nominal close. These are operationally analogous to claim-development triangles in casualty reserving and can be tracked with the same chain-ladder machinery. The volume is small in any single period but accumulates over the cohort lifecycle, particularly for agentic products with long-running workflows.

\paragraph{Endogenous churn at the cap.} A user who hits a hard cap mid-period faces a service halt, which converts into a non-trivial probability of cancellation, downgrade, or substitution to a competing product or to direct API access at the next renewal opportunity. We refer to this as \emph{endogenous churn at the cap}, structurally analogous to lapse on the insurance side except that the trigger is the binding event itself rather than a separate financial decision. Operationally, the relevant object is a hazard rate for user $i$ that depends on cap-hit history:
\begin{equation}\label{eq:churn-hazard}
h_i(t) \;=\; h_0(t) \cdot \exp\!\Big(\beta_1 \cdot \mathbb{1}[U_i(t) \geq K_i] + \beta_2 \cdot \mathrm{HitCount}_i(t)\Big),
\end{equation}
fitted via standard survival-analysis machinery (Cox or Weibull AFT in \texttt{lifelines}), where $\mathrm{HitCount}_i(t)$ is the cumulative number of cap-hits user $i$ has experienced up to period $t$. The two coefficients have distinct operational interpretations: $\beta_1 > 0$ captures the post-hit hazard step that persists while the user remains in the binding region; $\beta_2 > 0$ captures the per-additional-hit incremental fatigue accumulated across periods. The reserve calculation conditioned on a fixed portfolio understates this dynamic; a defensible practice is to refit the effective portfolio size $n_t = \sum_i \mathbb{1}[\text{user $i$ still subscribed at $t$}]$ at each reset using the fitted hazard, treating both $F_N, F_S$ and the portfolio composition as evolving jointly. Heavy users who hit the cap are over-represented among the lost subscribers, which shifts $F_N, F_S$ downward in the retained portfolio and shifts premium income downward in absolute terms. This is the operational problem Anthropic faced through the post-August-2025 period and is likely the single most important behavioral correction to standard reserve calculations for hard-cap LLM contracts.

\paragraph{Relation to mechanism-design pricing.} The framework here operates downstream of menu design. \citet{bergemann2026menu} derive the optimal $(P_i, K_i)$ menu under a risk-neutral buyer and full information about the buyer's value distribution; our framework takes any committed menu and computes $R$ and the operational metrics. The two are not substitutes. A capped-usage SaaS provider needs both: a mechanism-design step to set $(P_i, K_i)$ given the seller's beliefs about the user-value distribution, and a portfolio actuarial step to determine the reserve $R$ and to monitor whether the chosen $(P_i, K_i)$ remain solvent under realized consumption. The reserve calculation is the feedback signal that triggers re-running the mechanism-design step when the consumption distribution shifts.

\subsection{Behavioral component}\label{sec:behavioral}

The most distinctive feature of capped-usage contracts is the within-period behavioral response of users to the proximity of the cap and the reset. Both the cap $K_i$ in capped-usage AI and the out-of-pocket maximum (OOPM) in health insurance are kinks in the user-facing price schedule: a discontinuity in the marginal price at a threshold of cumulative consumption. The location of the kink relative to the price axis differs across domains---in health insurance, the marginal price \emph{drops} at the OOPM (the insurer absorbs further claims); in capped-usage AI, the marginal price \emph{jumps} at $K_i$ (the user absorbs further consumption). But the kink itself produces the same intertemporal substitution response: rational users accelerate consumption to capture value on the favorable side of the discontinuity, and the response intensifies as the reset approaches.

The intertemporal substitution under kinked health contracts is well documented, with care needed in attribution. \citet{manning1987rand} established the static price elasticity of medical demand using coinsurance variation in the RAND Health Insurance Experiment; this is the foundational evidence of price-sensitive medical demand, but does not isolate the dynamic component near a OOPM-style threshold. \citet{brotgoldberg2017deductible} document the converse end of the contract: spending under a deductible falls by 42\% with the asymmetric response concentrated in elective and preventive care. The dynamic response near a OOPM-style threshold is identified in \citet{einav2015response} and \citet{arondine2015moral}, using Medicare Part D and employer-plan nonlinear contracts respectively. \citet{diazcampo2022dynamic} formalizes the dynamic component in a structural model, finding that approximately 40\% of total moral-hazard spending is dynamic---driven by intertemporal substitution toward periods of low marginal price---rather than static price sensitivity.

In the capped-usage AI case the kink is at $K_i$ and the reset is at $T$. Two behavioral consequences follow. First, front-loading before the cap binds: as $U_i(t) \to K_i$, the user faces a discontinuous rise in marginal price and accelerates remaining elective demand before the discontinuity. Second, pull-forward before the reset: as $T - t \to 0$ with $U_i(t) < K_i$, the user accelerates consumption to avoid wasting the remaining non-fungible allowance, the dual of OOPM acceleration in the health case. Anthropic's introduction of weekly limits in August 2025 is, in this language, a non-trivial change in the price schedule: from a flat marginal cost of zero across the period to a kinked schedule with a hard threshold. The published rationale invoked the same risk-pooling logic (``usage is pooled across the org'') that insurers use for community rating, and the predicted behavioral response---concentration of consumption among power users in the days before reset, contraction of consumption in the days before $K_i$ binds---is identical in structure to the OOPM literature even though the direction of the price discontinuity is opposite.

Three modeling features follow. (i)~Time-to-reset $T - t$ should enter the frequency intensity as a covariate: $\lambda_i(t) = \lambda_i^{0} \cdot g(T - t, U_i(t))$, where $U_i(t)$ is the cumulative consumption to date. (ii)~The state $U_i(t)$ should enter the severity choice (model class, tool aggressiveness) as well as frequency. (iii)~The transition at $U_i(t) = K_i$ is the analogue of OOPM in health, and the behavioral discontinuity at the reset is the analogue of fiscal-year-end FSA spending. None of these features has been empirically modeled in the SaaS or AI subscription literature to our knowledge, and the corresponding identifying variation (weekly-limit introduction, tier-switching events) exists in firm-internal data even if not in public datasets.

Three testable predictions follow from this structure. \textbf{P1.}~Consumption patterns in Claude Code and ChatGPT should exhibit a post-reset bump driven by sunk-cost reasoning analogous to FSA end-of-year spending, detectable in aggregated usage logs. \textbf{P2.}~Cross-segment elasticity of demand at the marginal price should mirror \citet{manning1987rand} cross-segment patterns, with heavy users more inelastic than light users. \textbf{P3.}~The August 2025 introduction of weekly limits at Anthropic should have reduced aggregate consumption among power users by a magnitude estimable from observational telemetry, with the caveat that users above and below the new cap have materially different pre-period distributions; identification requires either difference-in-differences with explicit modeling of differential pre-trends (e.g., synthetic-control across user-week panels) or a regression-discontinuity design at the cap threshold. \textbf{P4.}~Higher-priced tiers should exhibit adverse selection identifiable as a wedge between expected loss in the chosen tier and expected loss predicted from demographics alone. We do not test these here; the framework merely renders them well-posed.

\subsection{Catastrophe risk}\label{sec:catastrophe}

Independence across users $C_i$---the working assumption in Section~\ref{sec:structure}---fails when a shared exogenous shock affects severity or frequency simultaneously across the portfolio. Three concrete catastrophe scenarios are relevant for capped-usage AI. \textbf{Model degradation}: a regression in model quality (post-training, alignment, or capability) increases token consumption per task across the user base. \textbf{Prompt-injection virality}: a discovered exploit propagates through tool-chains and inflates tool-call severity at the population level. \textbf{Liability-transfer breach}: in SaaS products with embedded liability transfer (the third example in Section~\ref{sec:mapping}), a single model failure mode can trigger correlated chargebacks across all contracts simultaneously, with severity bounded only by the embedded exposure cap.

The actuarial machinery for these risks is mature: Probable Maximum Loss (PML) calculations bound the exposure under worst-case correlated draws; copula-based dependence modeling (Gaussian, $t$, or Archimedean copulas) preserves marginal severity distributions while inducing tail dependence; catastrophe bonds \citep{mastroeni2022pricing} provide an alternative risk-transfer instrument when reinsurance is unavailable. The cyber-insurance literature provides templates for marked point process models of correlated incidents \citep{zeller2021comprehensive}, though that literature treats failure of the insured's own infrastructure rather than over-consumption by the insured's end-users; the modeling tools transfer with cosmetic changes.

The open problem is calibration. There are no historical loss data on AI catastrophe events at the frequency required to fit a tail-dependence parameter directly; analogue calibration from cloud outages, software supply-chain incidents, or model-rollout regressions is the only path.

\subsection{Worked example (Monte Carlo)}\label{sec:worked}

We illustrate the gap between naive unit-economics modeling and frequency--severity modeling with a synthetic Claude Code Max~20x-like portfolio: $n = 10{,}000$ subscribers, premium \$50/week each (\$500{,}000 portfolio income per period), non-fungible weekly cap $K = \$1{,}000$ per user. Parameters are illustrative orders of magnitude consistent with public Anthropic and OpenAI pricing; full calibration and code are in \texttt{sims/calibration.py}.

We compute portfolio aggregate cost $L = \sum_i C_i$ over $2{,}000$ Monte Carlo replications under three models, calibrated to the same expected per-user uncapped cost of \$30/week so that any difference in tail behavior reflects distributional shape rather than mean. \textbf{Naive baseline}: deterministic $\mathbb{E}[L] = n \cdot \bar\lambda \cdot \bar S = \$300{,}000$ with no variance, which is what a per-token cost-times-volume calculation produces. \textbf{Poisson--Gamma} (thin tail): $N_i \sim \mathrm{Poisson}(\lambda = 5)$ events/week, $S_{ij} \sim \mathrm{Gamma}(\alpha = 2, \theta = 3)$, mean severity \$6. \textbf{NB--LogNormal} (heavy tail at matched mean): $N_i \sim \mathrm{NB}(r = 1, p = 0.07)$ (mean $\approx 13$, var/mean $\approx 14$), $\log S_{ij} \sim \mathcal{N}(-0.311, 1.5^2)$, mean severity \$2.256 with severity standard-deviation-to-mean ratio of $\approx 2.9$.

\begin{table}[t]
\centering
\small
\begin{tabular}{@{}lrrrrr@{}}
\toprule
\textbf{Model} & $\mathbb{E}[L]$ & $\mathrm{VaR}_{0.99}(L)$ & $\mathrm{TVaR}_{0.99}(L)$ & \textbf{Loss ratio} & \textbf{Margin over $\mathrm{TVaR}$} \\
\midrule
Naive baseline & \$300{,}000 & \$300{,}000 & \$300{,}000 & 60.0\% & \$200{,}000 \\
Poisson--Gamma (capped) & \$300{,}019 & \$304{,}045 & \$304{,}641 & 60.0\% & \$195{,}359 \\
NB--LogNormal (capped) & \$299{,}729 & \$309{,}685 & \$310{,}984 & 59.9\% & \$189{,}016 \\
\bottomrule
\end{tabular}
\caption{\emph{Baseline actuarial illustration.} Portfolio aggregate cost statistics under three models at matched expected per-user cost. Premium income \$500{,}000 per period. Naive baseline ignores variance entirely; both compound models price the same expected loss but differ in tail behavior. This is a baseline scenario in which premium income comfortably covers expected loss; the heavy-tail effect is modest (a few percent) precisely because the cap and portfolio diversification absorb the per-user tail. Generated by \texttt{sims/tables/tab\_reserve\_comparison.py}.}
\label{tab:reserve}
\end{table}

Three readings of Table~\ref{tab:reserve} matter. First, the naive baseline understates tail risk by construction: it reports a \$200{,}000 margin over expected loss while either compound model gives a tighter \$189{,}000--\$195{,}000 margin over the 99th-percentile aggregate loss, a 2--6\% erosion of the operational cushion that is invisible to per-token analytics. Second, at matched expected loss, switching from Poisson--Gamma (thin tail) to NB--LogNormal (heavy tail) raises $\mathrm{VaR}_{0.99}$ and $\mathrm{TVaR}_{0.99}$ by approximately \$6{,}000 in this scenario, a modest 2\% effect at $n = 10{,}000$ subscribers because the cap is doing its work: per-user exposure is bounded at $K = \$1{,}000$, which absorbs the heavy individual tail before it aggregates to the portfolio. Third, this absorption is portfolio-size-dependent. Figure~\ref{fig:reserve} shows that per-user tail capital under NB--LogNormal decays approximately as $n^{-1/2}$ across $n \in \{500, \ldots, 100{,}000\}$, so the same family produces 5--10\% $\mathrm{TVaR}/\mathbb{E}[L]$ ratios at $n = 500$ and converges toward Poisson--Gamma at $n = 100{,}000$. Distributional choice matters more at small portfolio and looser cap.

\begin{figure}[t]
\centering
\includegraphics[width=0.85\linewidth]{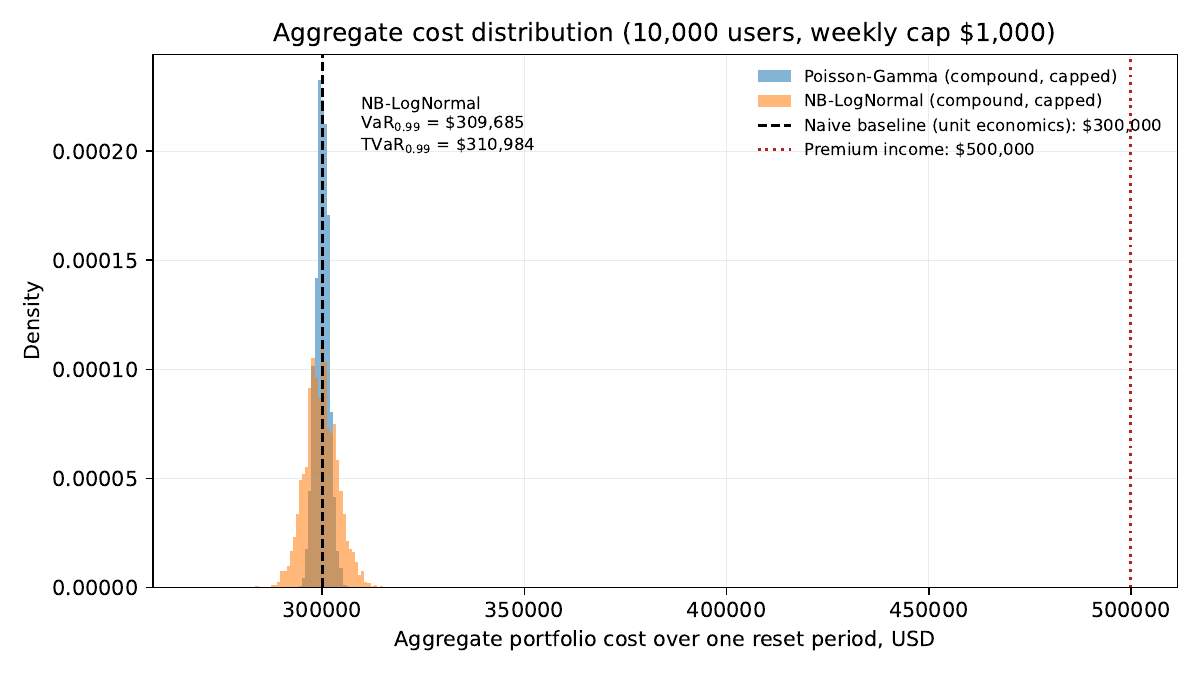}
\caption{Portfolio loss density for the two compound models under matched expected loss, with the naive point estimate and the premium income line for reference. Generated by \texttt{sims/figures/fig\_aggregate\_cost\_distribution.py}.}
\label{fig:agg}
\end{figure}

\begin{figure}[t]
\centering
\includegraphics[width=0.75\linewidth]{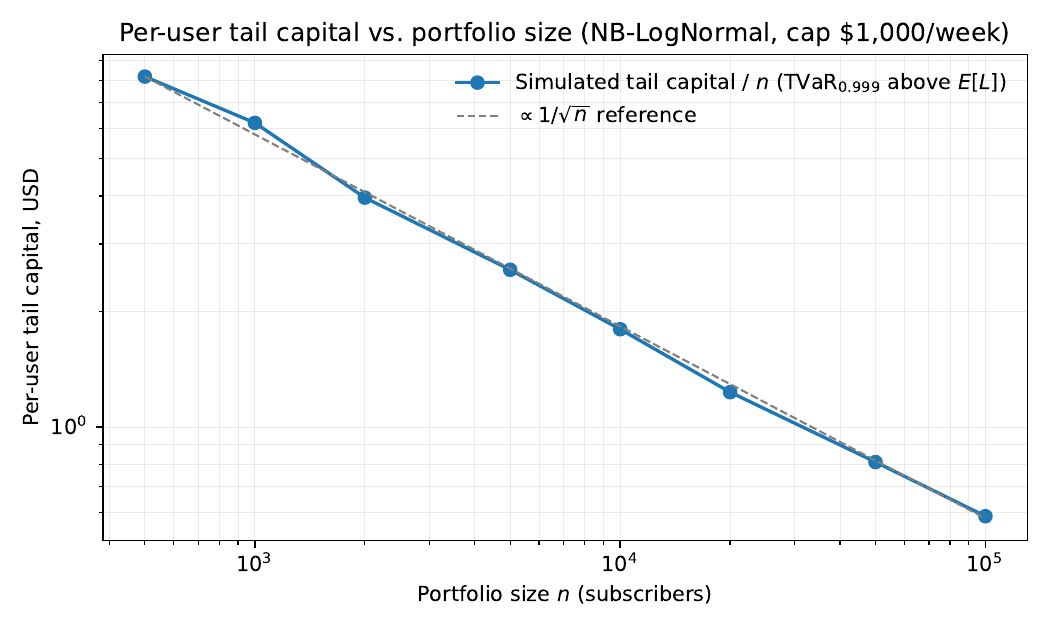}
\caption{Per-user tail capital ($\mathrm{TVaR}_{0.999}(L) - \mathbb{E}[L]$ divided by $n$) under NB--LogNormal, plotted against portfolio size with a $1/\sqrt{n}$ reference. Generated by \texttt{sims/figures/fig\_reserve\_by\_portfolio\_size.py}.}
\label{fig:reserve}
\end{figure}

The headline operational implication is that naive per-token economics overstates the comfort of an apparent margin by ignoring the variance that any reasonable distributional fit will surface, and that the distributional family choice continues to matter at the tail even when the cap is the dominant per-user constraint. The companion repository at \texttt{sims/} is reproducible from \texttt{uv sync \&\& uv run python ...}; all parameters are stated.

\subsection{What the cap is buying for the seller}\label{sec:policy}

The baseline scenario above (Table~\ref{tab:reserve}) is calibrated so that premium income comfortably covers expected loss; it represents a healthy operating point in which the cap is doing its work and the heavy-tail effect is modest. We now consider a deliberately extreme stress scenario: a population of heavy consumers whose expected per-user uncapped consumption is \$760/week, $15\times$ the \$50/week premium, with a NB--LogNormal model with $r = 2$, $p = 0.05$, $\mu = 0.996$, $\sigma = 2.0$ producing a long right tail. This represents the regime that motivated Anthropic's introduction of weekly limits in August 2025: a non-trivial share of users consuming materially more than baseline pricing would support if unconstrained. We hold the user behavior model fixed and compare four seller-side policy regimes, all on $n = 10{,}000$ subscribers.

\begin{table}[t]
\centering
\small
\begin{tabular}{@{}lrrrrr@{}}
\toprule
\textbf{Policy} & $\mathbb{E}[L]$ & $\mathrm{TVaR}_{0.99}(L)$ & \textbf{Loss ratio} & \textbf{Cap-hit} & \textbf{Margin} \\
\midrule
P0.\ Hard cap $K = \$1{,}000$ & \$5.4M & \$5.5M & 1079\% & 24.0\% & $-$\$5.0M \\
P1.\ Loose cap $K = \$5{,}000$ & \$7.4M & \$7.6M & 1481\% & 0.6\% & $-$\$7.1M \\
P2.\ No cap (flat-rate) & \$7.6M & \$7.9M & 1520\% & 0\% & $-$\$7.4M \\
P3.\ Pure pay-per-use, $P = 0$ & \$7.6M & \$8.0M & n/a & 0\% & n/a \\
\bottomrule
\end{tabular}
\caption{\emph{Stress scenario.} Counterfactual policy comparison for a heavy-consumer population whose expected uncapped consumption (\$760/week) is fifteen times the premium (\$50/week). All quantities are reported at retail pricing ($\kappa = 1$ in the notation of Section~\ref{sec:cost-vs-retail}); the seller's actual marginal cost is private and lower, so the loss-ratio magnitudes here are upper bounds rather than forecasts. Under sufficiently low $\kappa$ (roughly below $0.07$ for this calibration), even the no-cap P2 reaches expected-loss break-even, but the qualitative ordering across P0--P3 is preserved across all $\kappa \in (0, 1]$. This calibration is deliberately extreme, chosen to render the cap's effect visible and to mimic the regime that motivated Anthropic's August 2025 weekly-limit introduction. It is not a baseline representative scenario; the corresponding baseline is Table~\ref{tab:reserve}. Generated by \texttt{sims/studies/study\_policy\_alternatives.py}.}
\label{tab:policy}
\end{table}

Four observations follow. \textbf{First}, the no-cap regime (P2) is structurally insolvent for this population at the stated premium: realized aggregate loss exceeds premium income by an order of magnitude. This is the regime Claude Code Pro occupied before August 2025, and the regime that any subscription LLM tier without a hard cap occupies as long as the heavy-user tail is in the population. \textbf{Second}, a hard cap at the right level (P0) does not make the contract solvent in this population; it captures approximately \$2M of aggregate exposure ($\mathbb{E}[L]$ drops from \$7.4M to \$5.4M) by truncating individual exposures at \$1{,}000, at the cost of 24\% of users hitting the wall mid-period. The cap converts an unbounded liability into a bounded one, which is what allows the residual gap (premium repricing, tier redesign, model substitution) to be addressed by ordinary operational means. \textbf{Third}, a loose cap (P1) almost does no work in this population: it binds for 0.6\% of users and the aggregate is indistinguishable from no-cap. The choice of $K$ relative to the consumption distribution matters as much as the presence of a cap. \textbf{Fourth}, pure pay-per-use (P3) is solvent by construction---the user pays exact realized cost---and is exactly the API-direct model that LLM providers offer alongside subscription tiers. The trade-off the user makes when choosing P0 over P3 is paying for predictability rather than for risk transfer: in P0 the user accepts a hard service halt at $K$ in exchange for a fixed bill; in P3 the user accepts variable billing in exchange for unbounded service availability.

This decomposition is what the actuarial vocabulary makes visible. Without the cap, the seller absorbs the entire heavy tail; with the cap, the seller absorbs the body up to $K$ per user and the user absorbs the tail above $K$, exactly as in a policy-limit insurance contract. The cap is a retention point in the contractual sense: the seller's retained exposure is $\min(K, S_i^{\mathrm{agg}})$ and the user's residual exposure is the rest. The August 2025 weekly-limit change is, in this framing, the seller adjusting the retention point under a population shift in the heavy-tail mass, not a unilateral price change.

\section{Practical Implementation Notes}\label{sec:notes}

We collect here the vocabulary used throughout the paper, the operational artifacts a pricing or capacity team would track, and the tooling that supports them.

\subsection{Vocabulary}

Table~\ref{tab:vocab} maps actuarial terms to their capped-usage SaaS analogs. The mapping is one-to-one in all cases relevant to the framework. A team adopting the vocabulary does not need to learn new objects---only new names for objects they already manage.

\begin{table}[t]
\centering
\small
\begin{tabular}{@{}p{3.0cm}p{4.2cm}p{6.8cm}@{}}
\toprule
\textbf{Actuarial term} & \textbf{Capped-usage SaaS analog} & \textbf{Operational meaning} \\
\midrule
Premium $P_i$ & Subscription fee & Fixed income per subscriber per reset period \\
Frequency $N_i$ & Events/sessions/calls per subscriber & Count process per period \\
Severity $S_{ij}$ & Cost per event & Per-event monetary cost (tokens $\times$ model multiplier, check-ins $\times$ per-visit fee) \\
Aggregate loss $C_i$ & Realized period cost per subscriber & $C_i = \min(K_i, \sum_j S_{ij})$ \\
Policy limit / retention $K_i$ & The cap & Seller-side ceiling on per-subscriber exposure \\
Ceding & Tail above $K_i$ absorbed by the user & Contractually transferred residual exposure \\
Loss reserve $R$ & Capital held against tail aggregate cost & $R = \mathrm{TVaR}_{1-\alpha}(L) - \mathbb{E}[L]$ in this framework \\
IBNR & Events incurred but not yet billed at close & Lagged or in-flight events at the period boundary \\
Loss ratio & $\mathbb{E}[L] / \mathrm{premium~income}$ & Headline profitability indicator per cohort \\
Loading & Safety margin above expected loss in premium & Buffer for variance, capital cost, and profit \\
Credibility & Bayesian shrinkage in user-level fit & Weight on user-specific vs.\ portfolio-level estimate \\
Ratemaking & Tier design and repricing & Setting $P_i$ to match expected loss + loading + reserve cost \\
Catastrophe risk & Correlated tail event across portfolio & Model regression, viral exploit, identity model break \\
Coverage / exposure & Active subscriber base & Number of policies in force \\
\bottomrule
\end{tabular}
\caption{Actuarial vocabulary and SaaS analogs. A pricing team that already manages MRR, ARR, cohort retention, and unit economics has the underlying data; what changes is the model that consumes it.}
\label{tab:vocab}
\end{table}

\subsection{Artifacts to operate}

\paragraph{Metrics.} Loss ratio (realized $C_i$ aggregated to portfolio $L$ divided by premium income) is the headline indicator and should be tracked by cohort, tier, and acquisition channel. Frequency rate ($\bar N_i$ per period, by tier) and severity by segment provide diagnostic decomposition. Reserve coverage ratio (held reserve divided by $\mathrm{VaR}_{1-\alpha}(L) - \mathbb{E}[L]$) measures whether~\eqref{eq:reserve} is satisfied empirically.

\paragraph{Reporting cadence.} Monthly close on observed $C_i$ with same-period reserve adequacy check; quarterly reserve review with parameter refit and reserve resizing. The reset cadence of the contract ($T$ weekly or monthly) determines the close frequency; reserve refit at a slower cadence avoids over-reacting to single-period noise.

\paragraph{Tooling for the Python ML stack.} The actuarial reference implementations are in R (\texttt{actuar} for compound distributions and premium calculation; \texttt{ChainLadder} for IBNR-style reserving) and in Python for life insurance (\texttt{lifelib} for cohort projection, \texttt{chainladder-python} for reserving). For a SaaS modeling team operating on a standard Python ML stack (\texttt{numpy}, \texttt{pandas}, \texttt{scipy.stats}, \texttt{statsmodels}, \texttt{scikit-learn}, \texttt{pytorch}), the actuarial-classical libraries are mostly vestigial: \texttt{lifelib} solves life-insurance projection problems, not non-life compound loss, and \texttt{actuar} requires bringing R into the production stack. The minimal Python implementation of the framework requires only: \texttt{scipy.stats} (NB, LogNormal, Gamma, GPD samplers and densities), \texttt{statsmodels} (Tweedie GLM family for closed-form Poisson--Gamma compound fitting), \texttt{lifelines} (Tobit-style censored MLE via survival-analysis machinery), and one of \texttt{numpyro} or \texttt{PyMC} for hierarchical Bayesian update of $(F_N, F_S)$ priors. The companion \texttt{sims/} repository uses only \texttt{scipy}, \texttt{numpy}, \texttt{pandas}, and \texttt{matplotlib}. \citet{bahl2025computation} surveys premium calculation principles in a cloud-insurance context; \citet{wang2022warranty} is the closest M\&SOM precedent for portfolio-level reserve management in a non-insurance service contract using funds-pooling for warranty exposure.

\paragraph{Cap-aware data engineering: where the implementation actually breaks.} Standard usage telemetry records $(\mathrm{user}, \mathrm{event}, \mathrm{severity})$ tuples. Implementing Section~\ref{sec:modeling} requires augmenting this with $(K_i, T_i, U_i(t))$---the contract terms and running consumption state---to support both the censored likelihood of Section~\ref{sec:severity-bias} and the time-to-reset features of Section~\ref{sec:behavioral}. Three operational frictions determine whether the framework lands in production:

\textbf{State reconstruction.} The running cumulative consumption $U_i(t)$ at the moment of each event is generally not in the telemetry pipeline; it lives in the cap-enforcement service, which is rarely owned by the data team. Reconstructing $U_i(t)$ for historical events either requires re-processing event-level logs against a temporal join (expensive) or instrumenting the enforcement path to emit $U_i(t)$ alongside the event (a backend change). Neither is a modeling task.

\textbf{Cap function ambiguity.} Anthropic's weekly cap, for example, is denominated in proprietary ``usage units'' rather than tokens directly; modeling severity in dollars requires reconstructing the cap function from public pricing announcements and reverse-engineering the per-model weights, with measurement noise that propagates into censoring estimates. Plan-level mid-period changes (upgrade, downgrade, comp credits, enterprise overrides) make $K_i$ time-varying in a way the textbook Tobit does not cover.

\textbf{Non-stationary cost dynamics.} Model-cost shocks driven by inference-stack optimization, model substitution (Sonnet $\to$ Opus mix), or hardware-class shifts move $S_{ij}$ on quarterly timescales by amounts that exceed the within-quarter standard deviation of the fitted distribution. Reserve calculations therefore chase a non-stationary target; the rolling-window or Bayesian update of Section~\ref{sec:reserve} is the right machinery, but the practitioner should expect refit cadence to be the binding constraint, not parameter precision.

The framework's mathematics is settled. The friction in deploying it is concrete data-engineering and product-shock management, not abstract methodology---which is the standard story for any modeling discipline meeting a moving production system. Actuarial science and modern data-science / ML engineering are different professional communities, but the gap that matters here is between either of those and the contracts-and-billing engineering that owns the variables the framework wants to consume.

\section{Conclusion}\label{sec:conclusion}

Capped-usage SaaS is an insurance category by structure, whether or not it is insurance by law. The actuarial machinery for pricing it, holding reserves against it, and modeling user behavior under it has existed for decades for closely related problems. Adopting it does not require new theorems; it requires recognizing that the relevant theorems are not in the cs.LG canon. The four open problems we identified---cap-aware severity estimation, adverse selection across tiers, behavior near reset, and correlated catastrophe risk---are each tractable with existing tools, and each currently being solved by gut feel in production across LLM services, cloud platforms, corporate benefit platforms, and liability-transfer products alike.

The capped-usage SaaS industry will continue to reinvent these tools in public. Anthropic's weekly limits, OpenAI's tier-specific model gating, Vercel's overage rate schedules, Cloudflare's tier redesigns, and the cascade of price changes through 2025 and 2026 are responses, mathematically, to the same reserving, overage-pricing, and adverse-selection problems that the actuarial literature has tools for, and that the cs.LG and economics-of-AI corpora currently do not cite. Routing a fraction of the modeling effort now spent on per-unit unit-economics toward frequency--severity decomposition and tail-aware reserving would compress that reinvention cycle considerably, and would free up modeling capacity for the problems that are genuinely new---induced demand near reset, correlated catastrophe across providers, fairness implications of community rating at the organization level---and that the existing literature does not yet cover.

\appendix

\section{Reproducible Code}\label{sec:appendix-a}

All Monte Carlo simulations and counterfactual studies are archived in the companion repository, citable via DOI \href{https://doi.org/10.5281/zenodo.20213155}{\texttt{10.5281/zenodo.20213155}}. Stack is Python $\geq 3.11$ under \texttt{uv} with \texttt{numpy}, \texttt{scipy}, \texttt{pandas}, \texttt{matplotlib}. All randomness flows through a single \texttt{numpy.random.Generator} seeded by $20260515$; sub-experiments derive deterministic sub-seeds.

Table~\ref{tab:repro} maps each figure and data table in the body to its generating script and output file.

\begin{table}[h]
\centering
\scriptsize
\setlength{\tabcolsep}{4pt}
\begin{tabular}{@{}p{4.2cm}p{5.4cm}p{5.2cm}@{}}
\toprule
\textbf{Body artifact} & \textbf{Generating script} & \textbf{Output file (in \texttt{output/})} \\
\midrule
Table 1 (\S\ref{sec:mapping}, mapping) & --- (provider pricing pages, May 2026) & --- \\
Table 2 (\S\ref{sec:vercel}, Vercel) & \texttt{studies/study\_vercel.py} & \texttt{tab\_vercel.csv} \\
Table 3 (\S\ref{sec:adverse}, mixed-pop) & \texttt{studies/study\_mixed\_population.py} & \texttt{tab\_mixed\_population.csv} \\
Table 4 (\S\ref{sec:worked}, baseline) & \texttt{tables/tab\_reserve\_comparison.py} & \texttt{tab\_reserve\_comparison.csv} \\
Table 5 (\S\ref{sec:policy}, stress) & \texttt{studies/study\_policy\_alternatives.py} & \texttt{tab\_policy\_alternatives.csv} \\
Table 6 (\S\ref{sec:notes}, vocabulary) & --- (paper text only) & --- \\
Figure 1 (\S\ref{sec:worked}) & \texttt{figures/fig\_aggregate\_\allowbreak{}cost\_distribution.py} & \texttt{fig\_aggregate\_\allowbreak{}cost\_distribution.pdf} \\
Figure 2 (\S\ref{sec:worked}) & \texttt{figures/fig\_reserve\_\allowbreak{}by\_portfolio\_size.py} & \texttt{fig\_reserve\_\allowbreak{}by\_portfolio\_size.pdf} \\
\S\ref{sec:severity-bias} numbers & \texttt{studies/study\_censoring\_bias.py} & (stdout report) \\
\bottomrule
\end{tabular}
\caption{Reproduction map: body artifact $\to$ generating script $\to$ output file. All scripts run from the repository root; all output files land in the \texttt{output/} subdirectory.}
\label{tab:repro}
\end{table}

To reproduce every artifact from a clean checkout:

To reproduce, fetch the archived snapshot from Zenodo (DOI \href{https://doi.org/10.5281/zenodo.20213155}{\texttt{10.5281/zenodo.20213155}}), extract, and run:

\begin{lstlisting}[language=bash]
uv sync
uv run python figures/fig_aggregate_cost_distribution.py
uv run python figures/fig_reserve_by_portfolio_size.py
uv run python studies/study_censoring_bias.py
uv run python studies/study_vercel.py
uv run python studies/study_mixed_population.py
uv run python studies/study_policy_alternatives.py
uv run python tables/tab_reserve_comparison.py
\end{lstlisting}

Calibration sources for the illustrative scenarios are public Anthropic, OpenAI, Vercel, Cloudflare, and Supabase pricing announcements as of May 2026, with all parameter choices documented in \texttt{src/saas\_actuaria/calibration.py} of the companion repository. No proprietary data is used or required.

\bibliographystyle{plainnat}
\bibliography{refs}

\end{document}